\documentclass{article}
\pdfoutput=1

\usepackage[preprint]{nips_2018}

\usepackage{pgffor}
\usepackage[utf8]{inputenc} 
\usepackage[T1]{fontenc}    
\usepackage{hyperref}       
\usepackage{url}            
\usepackage{booktabs}       
\usepackage{amsfonts}       
\usepackage{nicefrac}       
\usepackage{microtype}      

\usepackage{amsmath} 
\usepackage{amssymb}  
\usepackage{xspace} 
\usepackage{algorithm,algorithmic}

\usepackage{cite}
\usepackage{xcolor}
\usepackage{graphicx}

\usepackage{wrapfig}
\usepackage[font=small]{subcaption}
\usepackage{caption}
\usepackage{xspace} 

\usepackage{graphicx}
\usepackage{subcaption}

\DeclareCaptionFont{scriptsize}{\scriptsize}
\DeclareCaptionFont{tiny}{\tiny}

\newcommand{\lr}{learner\xspace}
\newcommand{\qs}{selector\xspace}
\newcommand{\tr}{teacher\xspace}
\newcommand{\daal}{DAAL\xspace}

\DeclareMathOperator*{\argmin}{arg\,min}
\DeclareMathOperator*{\argmax}{arg\,max}
\DeclareMathOperator*{\expectation}{\mathbb{E}}

\title{Distribution Aware Active Learning}

\author{Arash Mehrjou$^\dagger$\\
Department of Empirical Inference\\
Max Planck Institute for Intelligent Systems \\
\texttt{arash.mehrjou@tuebingen.mpg.de} \\
\And
Mehran Khodabandeh$^\dagger$ \\
School of Computing Science \\
Simon Fraser University \\
\texttt{mkhodaba@sfu.ca} \\
\And
Greg Mori \\
School of Computing Science \\
Simon Fraser University \\
\texttt{mori@cs.sfu.ca} \\
}

\begin{document}

\maketitle

\begin{abstract}
Discriminative learning machines often need a large set of labeled samples for training. Active learning (AL) settings assume that the learner has the freedom to ask an oracle to label its desired samples. Traditional AL algorithms heuristically choose query samples about which the current learner is uncertain. This strategy does not make good use of the structure of the dataset at hand and is prone to be misguided by outliers. To alleviate this problem, we propose to distill the structural information into a probabilistic generative model which acts as a \emph{teacher} in our model. The active \emph{learner} uses this information effectively at each cycle of active learning. The proposed method is generic and does not depend on the type of learner and teacher. We then suggest a query criterion for active learning that is aware of distribution of data and is more robust against outliers. Our method can be combined readily with several other query criteria for active learning. We provide the formulation and empirically show our idea via toy and real examples. 
\end{abstract}

\section{Introduction}
\label{Introduction}

{
\newenvironment{myquote}[1]%
  {\list{}{\leftmargin=#1\rightmargin=#1}\item[]}%
  {\endlist}%
\begin{myquote}{0.2in}
\vspace{-5pt}
\centering
\fontsize{9}{10}\selectfont{``\textsl{A rare pattern can contain an enormous amount of information, provided that it is closely linked to the structure of the problem space.''}
}
\\
\begin{flushright}
{\fontsize{7}{8}\selectfont{---Allen Newell and Herbert A. Simon , Human problem solving, 1972}} 
\end{flushright}
\end{myquote}}

Active learning algorithms need to choose the most effective unlabeled data points to label in order to improve the current classifier.  The core issue in designing an active learning algorithm is deciding on the notion of potential effectiveness of these data points.

At its heart, the intuition is the following.  Effective data points are those for which the current classifier is uncertain.  However, it is also crucial that these data points are common, in the sense that they come from high density regions of input space.  Such data points are likely to have a higher impact on future labeling and lead to better generalization.  Utilizing this structural information regarding the input space is important for active learning algorithms to focus labeling resources where it matters.

Merging structural information with AL has been studied before. However, due to the difficulty in capturing the manifold of high dimensional distributions, approximate unsupervised learning methods such as clustering have been used. Recent advances in generative models~\citep{goodfellow2014generative} especially variational likelihood-based~\citep{kingma2013auto} or those which give a scalar as an unnormalized density~\citep{rezende2015variational} have opened up a new window to efficiently and directly take unsupervised information into account. 

We take advantage of the recent progress in non-parametric density estimators to inform active learning algorithms about the structural information of the dataset. The structural information of the dataset is learned offline by a density estimator. This information is then combined with the query criterion of a conventional active learning process. 

\subsection{Contribution}
Our proposed method is simple and modular and can be combined with many existing AL algorithms. We show that this approach gives advantages in some current issues of active learning, i.e., {\it 1) robustness against outliers}, {\it 2) choosing batches  of data at each AL cycle}, and {\it 3)  biased initial labeled sets}. The method is explained in Sec.~\ref{sec:daal} and each of these motivations is discussed in Sec.~\ref{subsec:justification}. Each motivation is then supported by empirical results in Sec.~\ref{sec:experiments}.

\section{Related Work}
\label{subsec:related_works}
\textbf{Active learning (AL)} aims to ease the process of learning by interlacing training the classifier with data collection. To this end, it cleverly chooses the samples to be labeled by a trusted annotator called the \emph{oracle}. To ease the presentation, we assume two machines in this learning framework: \emph{\lr} and \emph{\qs}. The \lr is doing the main aimed classification task and the \qs chooses which sample to query. In addition, there exists an oracle who provides the true label for each sample asked by the \qs. Normally in AL, a small set of labeled samples $D_l$ is given alongside a large set of unlabeled samples $D_u$. We can describe active learning by a multi-cycle learning procedure. Initially, the \lr is trained by $D_l$ at cycle $t=0$. At each subsequent cycle, \qs decides which samples from $D_u$ must be labeled by the oracle. Then the knowledge of the \lr is updated by the new set of labeled samples and the procedure repeats until some criterion is met.

There are various strategies in AL based on the inputs/outputs of the \qs. Assume \qs implements the function $S:\mathcal{I}\to\mathcal{Q}$ with input space $\mathcal{I}$ and output space $\mathcal{Q}$. The output space is the same as the feature space of \lr since the query samples must be meaningful for the oracle. In the query-synthesis AL strategy, the \qs \emph{synthesizes} samples to query. These synthesized samples do not necessarily  belong to $D_u$. 

Recent advances in generative models especially likelihood-free approaches such as generative adversarial networks~\citep{goodfellow2014generative} have increased the interest in this strategy which used to be mostly of theoretical interest before~\citep{ZhuB17}. Another strategy is so called selective or \emph{stream-based} sampling where sampling from the data distribution $P(x)$ is free where \qs sees one sample at each learning cycle and decides whether to query the sample or discard it~\citep{smailovic2014stream}. Therefore, the input to the \qs function $S$ at cycle $t$ is a single sample $x_t\in D_u$. 

For many real-world tasks, a large set of labeled samples can be collected at once. This motivates the pool-based active learning where the input space of function $S$ is $D_u$ and \qs can choose the best query at cycle $t$ among the whole unlabeled samples ~\citep{lewis1994sequential}. We build our work upon the pool-based strategy, i.e., we assume the set of unlabeled samples $D_u$ is available beforehand. Once the \qs is given a single sample in stream-based or a set of samples in pool-based sampling, it needs a method to choose a sample to query. Assume that \lr approximates the conditional distribution $P(y|x)$ by $Q(\hat{y}|x;\theta(t))$ at cycle $t$ of the active learning process. The \qs at cycle $t$ chooses the query sample from $D_u$ based on looking at the current learner knowledge which is embedded in $\theta(t)$. The \qs may use $Q$ in many different ways and achieve different selection strategies. 

One popular method called \emph{uncertainty-sampling} chooses $x$ for which the classifier is least certain. There are several measures of uncertainty in the literature but the most used one is entropy $x_H^*=\argmax_x H(Y|x;\theta(t))$ ~\citep{shannon1948mathematical}.
Uncertainty sampling has pros and cons. Having intuitive interpretation and ease of implementation are among the positive properties. Moreover, it is modular and generic by seeing \lr as a black box which is only asked by the \qs for the confidence score for each sample. However, one major drawback of uncertainty-sampling is that the the decision of \qs at each cycle depends solely on $Q(y|x;\theta(t))$. Since $\theta(0)$ is learned by only a small set of initially labeled samples $D_l$, it may induce a significant bias in choosing the queries and consequently updating the \lr by biased samples. This condition may result in a myopic \lr that has a tiny chance to see samples from distant regions of the feature space if the initial labeled set does not contain samples from those regions~\citep{richards2011active}. The general idea of our work is to enhance the vision of \lr to give it a broader view of the feature space. 

The idea of using structural information in AL has been around and studied in previous work. Current work generally uses clustering as a method to encode structural information during active learning mainly because approximating $P(x)$ is not tractable. The idea behind~\citep{nguyen2004active} is to pre-cluster data and give more weight to the centroids of clusters as representative samples. In addition, repetitive labeling of the samples belonging to the same cluster is suppressed. This is an improvement over previous works~\citep{zhang2002active,zhang2000value,zhu2005kernel} which used cluster centroids as most interesting sample but did not provide any measure to avoid repeated sampling from the same cluster. Clustering based AL methods often have unrealistic assumptions about the distribution of data and also the distribution of the labels of samples withing each cluster. They often assume that data is well distributed into $\{1,2,\ldots,K\}$ clusters and once the cluster membership $k$ is known for a sample, its label is known as well. Another limitation is the simple distribution which is assumed for each cluster (e.g.\ Gaussian). 

Active learning and semi-supervised learning have been combined by various methods as an approach to take advantage of structural information. For example, similarity between unlabeled points are modeled as a graph with weighted edges in~\citep{ZhuB17} where a Gaussian field is constructed for which the generalization error can be efficiently approximated and used for query selection. However, the similarity measure between two samples is defined in terms of an RBF kernel which can be remote from reality. A semi-supervised active learning framework is proposed in~\citep{leng2013combining} to use class central samples as a guide to choose better class boundary samples. However, the full use of the data distribution is still missing. We propose a simple modular way to make use of an approximate data distribution that gives better query selection strategy.

There are a couple of recent works which also take structural information into account mainly in an indirect way. For example, ~\citep{bachman2017learning, ravi2018meta} uses meta-learning to learn an active learning strategy which can be transferred to other tasks. Similarly,~\citep{konyushkova2017learning} learns the AL strategy which is a regression method to predict the reduction in the classification error after labeling a sample and then transferring this strategy to novel tasks. Even though structural information has augmented traditional AL heuristics in these works, the explicit value of unnormalized probability has not been used. In addition the proposed methods are not modular and cannot be easily combined with previous AL strategies.

The cost of AL is mainly defined as the labeling cost which is proportional to the number of queried samples. However, this is not the only conceived cost. Retraining the classifier at each AL cycle also charges the user. ~\citep{shen2017deep} takes this cost into account and proposes a method to reduce it. Our proposed method reduces the retraining cost as well. It enables batch sampling while the samples of each batch are more informative than conventional batch sampling in AL.


The proposed method in this paper is modular, it can be combined with other AL strategies. Moreover, with minor modifications, it can be used with unconventional oracles that do not directly provide correct labels for the queried samples~\citep{xu2017noise, murugesan2017active}.

\vspace{-1mm}
\section{The proposed method: \daal}
\label{sec:daal}
The intuitive idea behind our work is that the \qs can make wiser decision in choosing queries if it has knowledge about the structure of the feature space fully represented in probability distribution $P(x)$ of data. However, normally we do not have direct access to distribution and are instead given a bunch of unlabeled samples generated from it represented by $D_u$. Hence, the first step is to approximate $P(x)$ by some function $Q(x;\psi)$ called \emph{\tr}. Structural information of $D_u$ is then distilled in $Q(x;\psi)$ and can be used to guide the \qs to choose more effective query samples. Because query selection at each active learning cycle is informed by distribution information of the dataset, we call our method Distribution Aware Active Learning (\daal). 
The information content of each sample is not fully determined only by the uncertainty of the current \lr about that sample. Formally speaking, uncertainty-sampling (US) defines the information content of sample $x$ with unknown label $y$, as ${\rm Info}_{\rm US}(x) = H(y|x;(\theta(t))$, where $P(y|x)\approx Q(\hat{y}|x;\theta(t))$. \daal augments the definition of the information content of sample $x$ with structural information $Q(x;\psi)$, i.e., $\Phi(x;\theta, \psi)={\rm Info}_{\rm DAAL}(x)$ is a function of $Q(x;\psi)$ and $ H(y|x;\theta(t))$. The detailed definition of $Q(x;\psi)$ and $\Phi(x;\theta, \psi)$ is discussed later in Section~\ref{sec:daal}. 

Here we provide the logic behind our idea and its formulation in its generic form. Assume we have a base criterion $\psi_b(x)$ for proposal of a conventional active learner. We suggest a new proposal criterion on top of this base criterion which is formulated as follows:

\begin{equation}
    x^* = \argmax_x \Phi(x;\theta,\psi)=\argmax_x \phi_b(x;\theta) \times [q(x; \psi)] ^ \beta.
    \label{eq:daal_query}
\end{equation}

In this formulation, $q(x; \psi)$ encapsulates the structural information of dataset $D_u$ which is already distilled in the \tr component. The hyper-parameter $\beta$ controls the $attention$ of the \qs to the \tr. Larger $\beta$ turns the \qs's decision for querying the samples more towards the knowledge of \tr than the current knowledge of the \lr. The criterion $\phi_b(x)$ can be any simple active learning criterion which is here assumed to be uncertainty sampling, i.e, it chooses samples which are closer to decision boundaries where the labels are most ambiguous. The major question now is how to distill the structural information of $D_u$ in the \tr component and use it to design $q(x;\psi)$. Next section presents one practical way to do so.



\subsection{VAE as density estimator for \daal}
\label{subsec:vae}
Variational autoencoder~\citep{kingma2013auto} is a setup for doing inference in a class of deep probabilistic models. The class of models can be almost any unsupervised density estimator with latent random variables. Here, we briefly present the essence of VAEs and show how it can be used in \daal.
As any other latent variable models of observed variables $x$, a new set of variables $z$ is introduced and the joint probability distribution over $\{x,z\}$ is factorized by Bayes formula $p(x,z)=p(x|z)p(z)$. The generative process is to first generate samples from prior distribution $z_i\sim p(z)$ and then generate samples from the conditional distribution $x_i\sim p(x|z_i)$. Inference means computing the conditional distribution $p(z|x)=p(x|z)p(z)/p(x)$ which requires computing the \emph{evidence} factor $p(x)$. 
The evidence is hard to compute because the marginalizing integral 
$\int_{z\in\mathcal{Z}}p(x)=\int{p(x|z)p(z)dz}$ 
is taken over exponentially many configurations of latent variables. 
Variational inference tries to approximate $p(z|x)$ with $q_\psi(z|x)$ chosen from a family of functions indexed by $\psi$. 
This can be done by minimizing the Kullback-Leibler divergence between these two distributions:

\begin{equation}
    q_{\psi^*}(z|x) = \argmin_\psi {\rm KL}(q_\psi(z|x) || p(z|x))
    \label{eq:posterior_KL}
\end{equation}


This criterion is hard to compute because of the presence of intractable $p(z|x)$. Algebraic re-arrangement of the terms reveals the following equations which are of our most interest

\begin{align}
    {\rm ELBO}(x;\psi) &= \expectation_q[\log p(x,z)] - \expectation_q[\log q_\psi(z|x)]\\
    \log p(x) &= {\rm ELBO}(x;\psi) + {\rm KL}(q_\psi(z|x) || p(z|x))\label{eq:logp}
\end{align}

Jensen's inequality ensures that KL-divergence is non-negative. This said and because the lefthand side of Eq.~\ref{eq:logp} does not depend on $\psi$, we can maximize ${\rm ELBO}(x;\psi)$ as an implicit way to minimize Eq.~\ref{eq:posterior_KL}. Therefore, we have ${\rm ELBO}(x;\psi)<\log p(x)$ and after the optimization is completed, ${\rm ELBO}(x;\psi^*) \approx \log p(x)$. Exponentiation both sides, we have an approximate to the probability distribution of observed data $\exp({\rm ELBO}(x;\psi^*))=q(x;\psi^*)\approx p(x)$. This approximation to $p(x)$ is then used in \daal to design the structural part of Eq.~\ref{eq:daal_query}. In practice, we observed that passing the value of $q(x;\psi^*)$ through a sigmoid function, i.e., $q(x;\psi^*)\leftarrow{\rm \sigma}(q(x;\psi^*))$ gives a better performance where $\sigma(x)=1/1+\exp(-x)$.

\subsection{Motivations for \daal}
\label{subsec:justification}

In this section, we investigate the motivations for \daal and how augmenting a base \qs with structural data can lead to improvements over traditional AL methods. The motivations are listed and described below.

{\it Robustness against outliers---} The uncertainty based criterion $\phi_b(x)$ only cares about the distance of samples from the decision boundary. There is always a chance that this large distance is caused by an outlier for which $P(x)\ll 1$. Choosing an outlier will misguide the \lr and changes the decision boundary dramatically. Query criterion of Eq.~\ref{eq:daal_query} on the other hand takes the relative rareness of outliers into account through its second term $[q(x;\theta)]^\beta$ and prevents \qs from mistakenly choosing them and asking the oracle for their labels. This not only saves the decision boundary from being affected by the outliers, but also removes the extra cost imposed on the oracle to label a useless sample.

A practical example of this is object detection, where obtaining bounding boxes of objects in images is a costly process. Active learning can decrease this cost by proposing bounding boxes that most likely contain an object to the oracle and only ask for the label. However, finding those bounding boxes is not an easy task and is prone to outliers (bounding boxes with no objects). This is where our method can be useful. This can be done by learning a density estimator that can predict the likelihood of bounding boxes containing objects. 

{\it Batch active learning---} uncertainty based sampling can choose only one sample at a time. The reason for this limitation is clear. If the decision function $y = f(x; \theta(t))$ of the classifier is slowly varying with respect to $x$, the entropy of the labels $H(y|x;\theta(t))$ and consequently $\phi_b(x;\theta)$ changes slowly as well:
\begin{equation}
|\phi_b(x_1;\theta) - \phi_b(x_2;\theta)| \to 0 \implies |x_1 - x_2| \to 0
\label{eq:based_criterion_continous}
\end{equation}
This implies that if a batch $B = \{x_1^*, x_2^*,\ldots\}$ of samples is chosen by uncertainty sampling instead of a single sample, the set $B$ lacks diversity and the information content of $B/{x_1^*}$ considerably decreases when $x_1^*$ is known to the \lr. This is a known effect that choosing the highest score samples from the AL pool gives samples with low diversity~\citep{guo2008discriminative}. \daal has an automatic means to mitigate this problem and enable \qs to choose multiple samples at each AL cycle. The reason for this higher diversity can be explained as follows. Assume the unlabeled samples of the set $D_u$ are sorted by the criterion $\Phi(x;\theta, \psi)$ of Eq.~\ref{eq:daal_query}. For two samples $\{x_1, x_2\}$ with close $\Phi$ scores, we can write:
\begin{align}
    \Phi(x_1;\theta,\psi)\approx\Phi(x_2;\theta, \psi) &\implies \phi_b(x_1;\theta) \times [q(x_1; \psi)] ^ \beta \approx \phi_b(x_2;\theta) \times [q(x_2; \psi)] ^ \beta\\
    &\implies \frac{\phi_b(x_1;\theta)}{\phi_b(x_2;\theta)} \approx \left[\frac{q(x_2; \psi)}{q(x_1; \psi)}\right]^\beta
\end{align}
For this equation to hold, ${x_1, x_2}$ does not need to be close to each other in the feature space. The left-hand side of the above equation can be far from unity for a multi-modal $q(x;\psi)$. This effect gets magnified for large values of $\beta$ resulting in more diversity in batch $B$ chosen by criterion $\Phi(x;\theta,\psi)$.

{\it Starting from scratch by annealing $\beta$ ($\infty\to\beta\to 0$)---} Many active learning algorithms depend on an initial set of labeled samples. This may bias the AL process towards small regions of the feature space especially when the initial set is small or unrepresentative of the underlying distribution. \daal deals with this problem with no need for manually selecting the initial set. Here we introduce a dynamical approach inspired by non-autonomous dynamical systems that changes the problem setting over time~\citep{bengio2009curriculum,mehrjou2018analysis,mehrjou2017annealed}. Assume that at the very beginning of AL process, $\beta$ is large and $\Phi(x;\theta, \psi)$ is mainly influenced by $q(x;\psi)$. This amounts to choosing samples from $D_u$ with highest values of $q(x;\theta)$ which are most representative samples from data distribution. For example, in a multi-modal distribution, this criterion ensures us that in the beginning, the most representative samples of each mode are selected. As AL proceeds, we decrease the value of $\beta$. This results in a more prominent role for uncertainty term of $\Phi(x;\theta,\psi)$ which is $\phi_b(x;\theta)$. That is, more focus on precising the decision boundaries in the regions of the feature space where there still exist ambiguity in terms of label uncertainty. Simply speaking, by annealing the attention hyper-parameter $\beta$ from $\infty$ (some large value) to $0$, in the beginning, the \qs is highly attentive to the \tr and selects a diverse set of representative samples. This results in finding coarse decision boundaries in the beginning. As $\beta$ decreases over AL cycles, \qs becomes less attentive to the teacher and more attentive to \lr. Being more attentive to the \lr means choosing more samples from regions of the feature space that help resolve ambiguity of the \lr in those regions.

In the next section, we provide experiments to empirically show the aforementioned points.

\newpage

\section{Experiment}
\label{sec:experiments}
\subsection{Toy example}

\begin{wrapfigure}[13]{r}{.25\textwidth}
 \captionsetup{width=0.97\linewidth}
\vspace{-2mm}
   \vspace{-\intextsep}
    \begin{minipage}{0.25\textwidth}
    \centering
    \includegraphics[width=\textwidth]{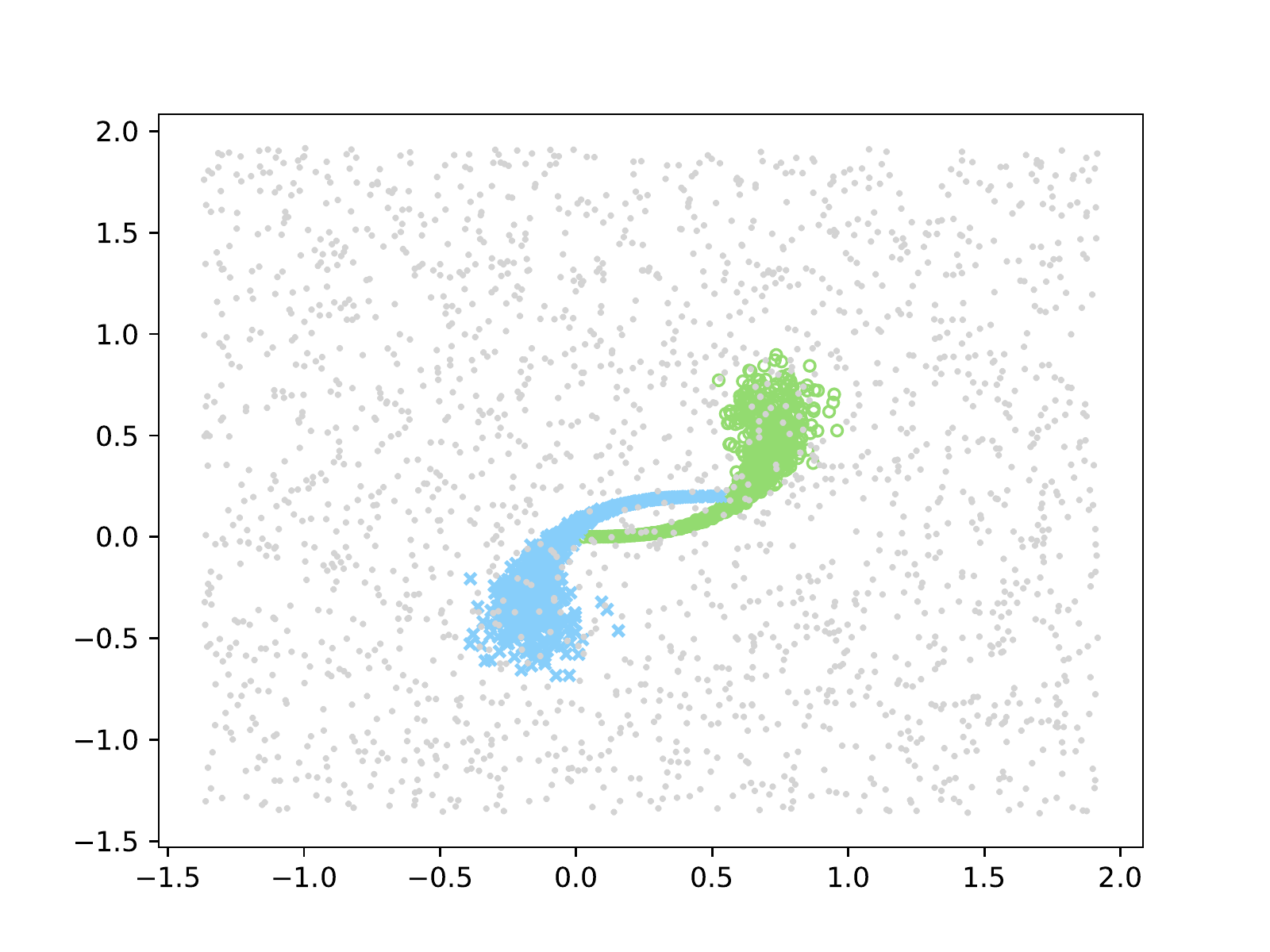}
    \centering

    \end{minipage}
     
\centering\vspace{-2mm}\caption{ Active learning data pool. Green and Blue are two classes of interest and the gray dots are outliers}\label{fig:toy_data_fig}
\end{wrapfigure}

We design a toy example to showcase the efficacy of our method and visualize its performance. Assume the task is to separate two classes. Class conditional densities from which the samples of each class are generated are represented by  $p(x|y=c_1)$ and $p(x|y=c_2)$. In addition, when active learner explores the world to query new samples, it may encounter outliers or noisy samples which come from neither \mbox{$p(x|y=c_1)$} nor $p(x|y=c_2)$. To simulate this effect, we assume a third class distribution called \emph{outlier} distribution which is represented by $p_{out}(x)$. The active learner sees samples in the world that come from either $p(x)=\sum_{y\in\{y_1,y_2\}}p(x|y)p(y)$ or $p_{out}(x)$. From the active learner's perspective, we assume the samples come from a single distribution represented by $p_{al}(x)$ which is itself a mixture of true data and outliers:
\begin{equation}
    p_{al}(x)=\mu_1p(x)+\mu_2p_{out}(x), \quad \mu_1+\mu_2 = 1
\end{equation}

The class conditional distributions for the following experiment are represented as colored dots in Fig.~\ref{fig:toy_data_fig}. We assume $p_{out}(x)$ is uniform over a bounding box around the domain of $p(x)$. A variational autoencoder is then trained on samples from $p(x)$ and the heatmap of $q(x;\psi)$ (see sec.~\ref{subsec:vae}) for different values of $\beta$ is depicted in Fig.~\ref{fig:heatmap_vae}.
The \qs uses eq.~\ref{eq:daal_query} to sort the pool of unlabeled samples and find the best ones to query. As stated in section~\ref{sec:daal}, \daal becomes normal active learning (blind to distribution) for $\beta=0$. As $\beta$ increases, the role of the distribution becomes prominent. We have chosen an intermediate value $\beta=0.8$ in this experiment. Fig.~\ref{fig:before_after} illustrates how the proposed sampling strategy of Eq.~\ref{eq:daal_query} influences a conventional active learning criterion (e.g. label entropy) at each AL cycle. Using Eq.~\ref{eq:daal_query}, the outliers that gain large values of $\phi_b(x)$ will get a lower overall score due to the structural term $q(x;\psi)$ and have lower chance of being queried by the \qs.

\begin{figure*}[!b]
    \centering
    \begin{minipage}{0.44\textwidth}
    \centering
    \foreach \n in {0.2, 0.4}
    {
    \begin{subfigure}[t]{0.45\textwidth}
        \centering
        \includegraphics[width=0.85\textwidth]{figures/p_x/px_map_\n.png}
        \subcaption{\small $\beta={\n}$}
    \end{subfigure}%
    }
    \par\vfill
    \centering
    \foreach \n in {0.8, 1.6}
    {
    \begin{subfigure}[t]{0.45\textwidth}
        \centering
        \includegraphics[width=0.85\textwidth]{figures/p_x/px_map_\n.png}
        \subcaption{$\beta={\n}$}
    \end{subfigure}%
    }
    \captionsetup{width=0.8\linewidth}
    \caption{\small Heatmap of $q(x;\psi)^\beta$ for different values of $\beta$. The function is more concentrated on higher values of the distribution for higher values of $\beta$. It can be seen as an inverse temperature parameter~\citep{de2013non}.}
    \label{fig:heatmap_vae}
    \end{minipage}
    \hfill
    \begin{minipage}{0.55\textwidth}
    \vspace{-6mm}
    \foreach \n in {0, 1, 2}
    {
    \begin{subfigure}[t]{0.32\textwidth}
        \centering
        \includegraphics[width=1.05\textwidth]{figures/before_after/before_after_step_\n.png}
        \vspace{-10mm}
        \subcaption{\small Cycle $\n$}
    \end{subfigure}%
    }
    \captionsetup{width=0.97\linewidth}
    \caption{\small Every column shows one cycle of active learning. Top row illustrates the values of the base criterion $\phi_b(x;\theta)$ for every sample $x$ in the pool. The black curve depicts the boundary of the classifier. In the bottom row, the samples are color coded by the scores that \daal gives to each sample based on Eq.~\ref{eq:daal_query}. Queried samples are shown with blue crosses.}
    \label{fig:before_after}
    \end{minipage}
    \vspace{-1mm}
\end{figure*}

\begin{figure*}[!h]
     \centering
     \begin{minipage}{0.69\textwidth}
     \foreach \n in {0, 1, 2}
     {
     \begin{subfigure}[t]{0.32\textwidth}
         \centering
         \includegraphics[width=1.1\textwidth]{figures/al_steps/beta=0/current_step_\n.png}
         \par\vfill
         \includegraphics[width=1.1\textwidth]{figures/al_steps/beta=0.6/current_step_\n.png}
         \subcaption{\small cycle $\n$}
     \end{subfigure}%
     }
     \captionsetup{width=0.9\linewidth}
     \caption{\small Every column illustrates the heatmap of unlabeled pool computed by our method (\textbf{bottom row}) and baseline (\textbf{top row}) in an AL cycle. Query points ($10$ points at each cycle) are shown with either blue crosses (outliers), or green circles (inliers). }
     \label{fig:toy_iterations}
     \end{minipage}
     \hfill
     \begin{minipage}{0.30\textwidth}
        \captionsetup{width=0.9\linewidth,justification=centering}
         \begin{subfigure}{0.95\textwidth}
        \includegraphics[width=\textwidth]{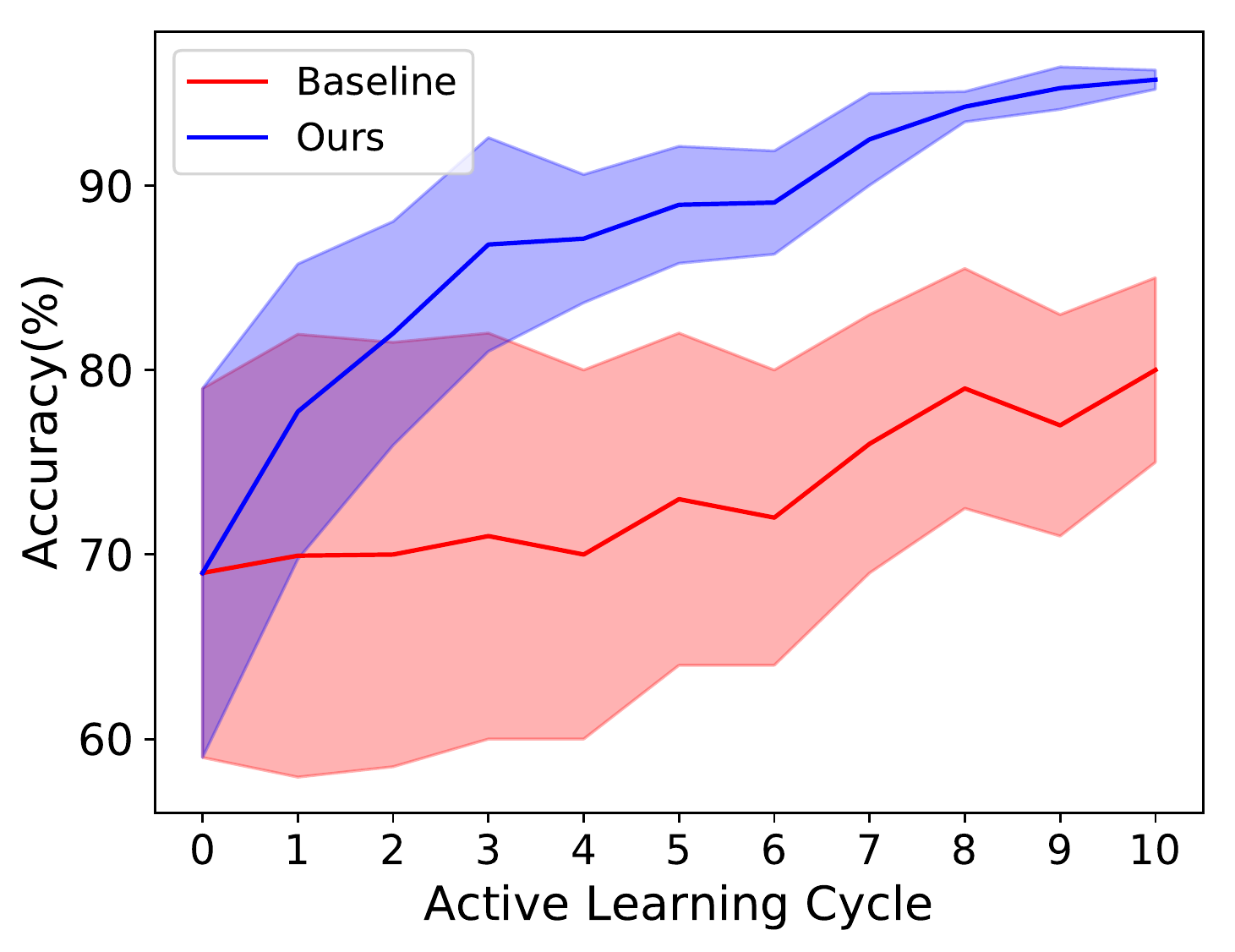}
        \subcaption{\small Accuracy}
        \label{fig:toy_accuracy}
        \end{subfigure}
        \par\vfill
        \begin{subfigure}{0.95\textwidth}
        \centering
        \includegraphics[width=\textwidth]{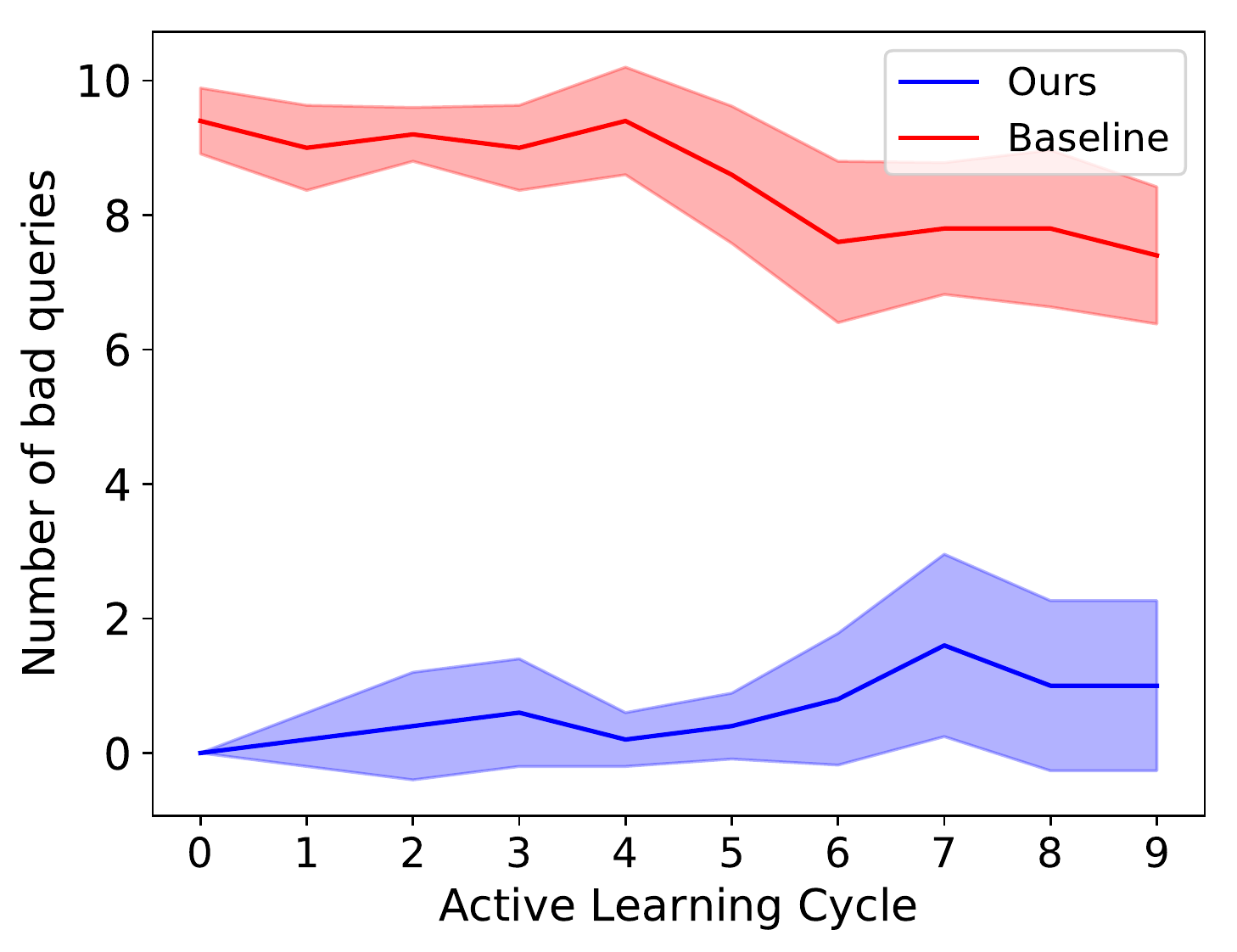}
        \subcaption{\small Number of outlier queries (lower the better).}
        \label{fig:toy_num_outlier}
    \end{subfigure}
\centering
\captionsetup{width=0.9\linewidth,justification=raggedright}
\caption{\small Quantitative results on toy example}\label{fig:toy_quantitative}
     \end{minipage}
 \end{figure*}

To show the effectiveness of our algorithm, we compare a normal active learning process ($\beta =0$) with our method ($\beta >0$, in this case $\beta=0.8$). At each iteration, the \qs queries $10$ data points. Qualitative results are shown in Fig.~\ref{fig:toy_iterations} and quantitative results are shown in Fig.~\ref{fig:toy_quantitative}. Fig.~\ref{fig:toy_accuracy} shows the average accuracy of the classifier for $10$ runs trained on queried samples at each iteration of active learning. At the beginning (cycle $0$) of each run, the labeled set (of size $2$) is initialized at random ($1$ sample per class). However, this initial labeled set is identical for both our method ($\beta=0.8$) and the baseline ($\beta=0$). In all the experiments performed on the toy example we used a simple multilayer perceptron with two hidden layers containing $8$ and $4$ nodes, respectively, with ReLU as activation. We used the VAE architecture of ~\citep{kingma2013auto} as the density estimator.


\subsection{High Dimensional Data}
\begin{figure*}[b]
    \centering
    \begin{subfigure}[t]{0.41\textwidth}
        \centering
        \includegraphics[width=\textwidth]{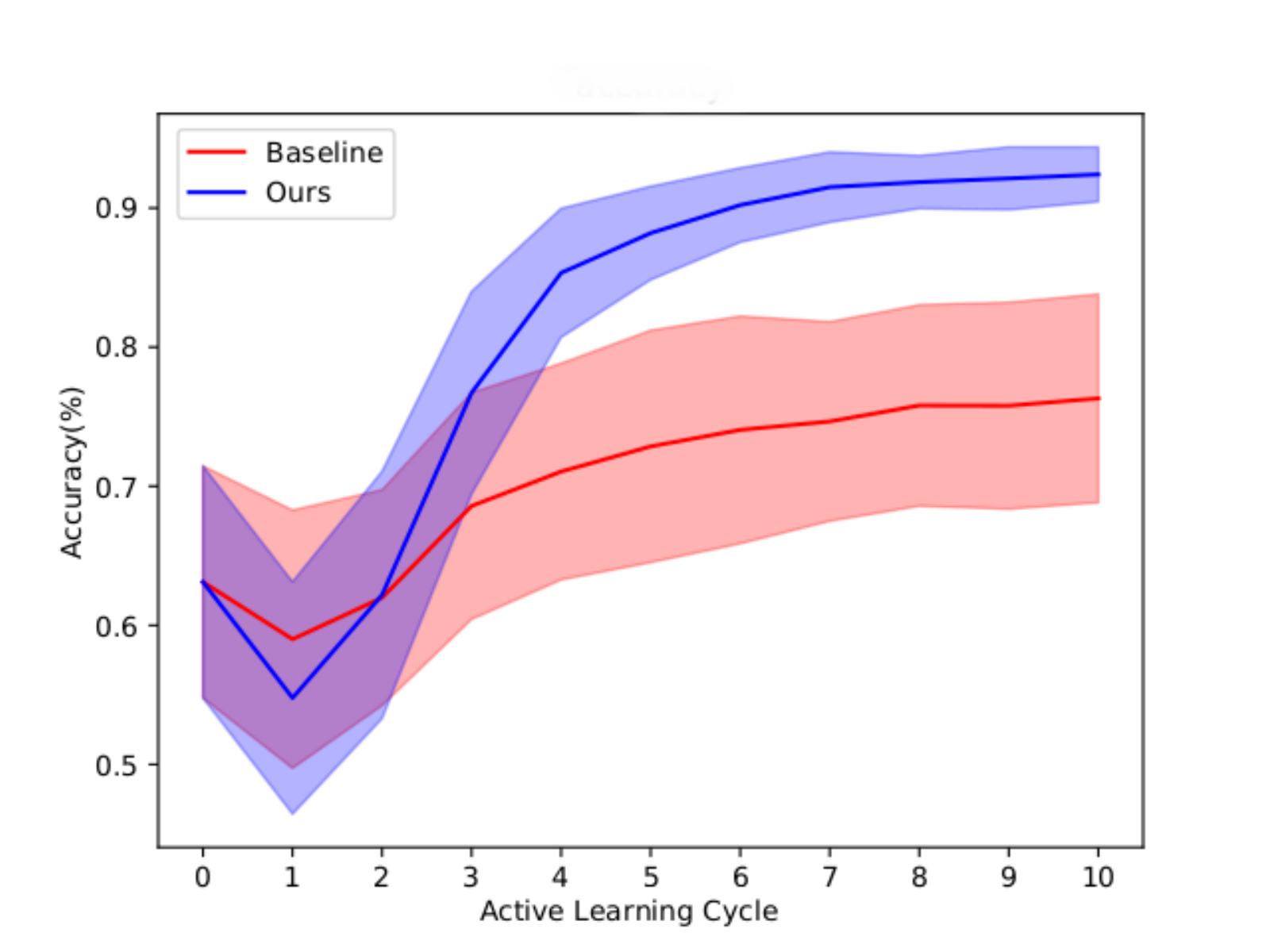}
        \subcaption{}
        \label{fig:mnist_accuracy}
    \end{subfigure}%
  \begin{subfigure}[t]{0.41\textwidth}
   \includegraphics[width=\textwidth]{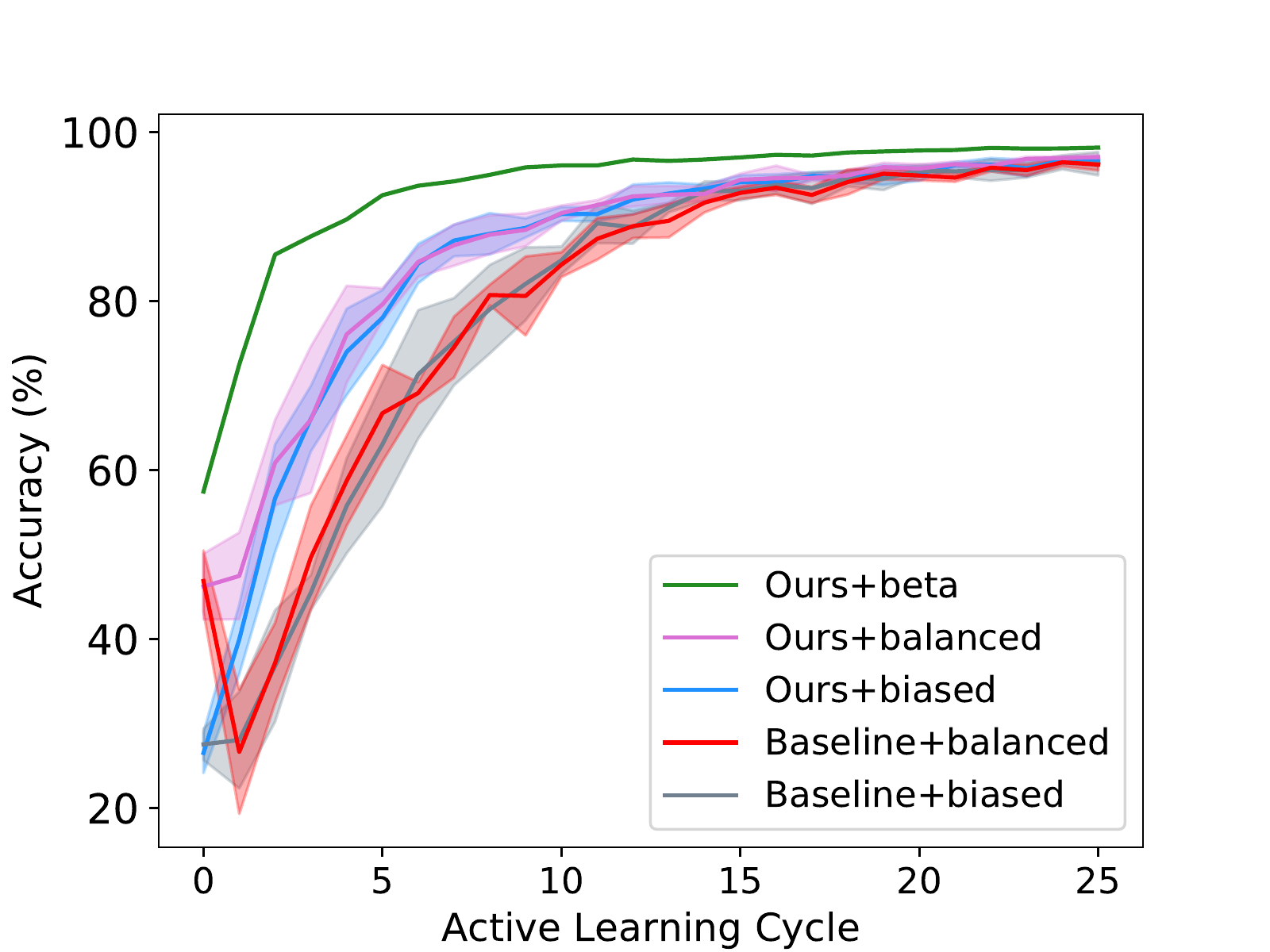}
   \caption{}
    \label{fig:anneal}
   \end{subfigure}
   
   
    \caption{\small Quantitative results on MNIST. (a) Comparison between our method and baseline in presence of outliers.
    (b) Comparison of performance of our annealing method between and the baseline in absence of outliers.
   }
    \label{fig:mnist_quantitative}
    \vspace{-1mm}
\end{figure*}

As a high dimensional example, we test \daal on MNIST, a dataset containing handwritten digits\citep{lecun1998mnist}. To mimic the notion of outlier and inliers, we use the first five digits $\mathcal{X}=\{0,1,2,3,4\}$ as the dataset of interest and the remaining digits as outliers. We first randomly sample $1000$ images of each digit in $\mathcal{X}$ from the training set (in total $5000$) to train the generative model, and use the rest of the training set combined with a portion of outliers as the \emph{pool} from which the \qs is to choose query samples (in our experiments the total number of outliers is two times more than inliers). The classifier is validated on the MNIST test set. Fig.~\ref{fig:mnist_accuracy} shows the performance of our method compared with the baseline. In all the experiments on MNIST, we used LeNet~\citep{lecun1998gradient} as the classifier and VAE~\citep{kingma2013auto} as the density estimator.



\subsection{Annealing $\beta$}

In the previous experiments, we showed the usefulness of \daal for a fixed value of $\beta>0$ as a measure that takes into account the structural information of a dataset. In this section, we investigate the effect of annealing $\beta$ from a large value at he beginning towards a small value as active learning proceeds. We show that \daal is not only robust against outliers, but also gives a better performance even in absence of outliers. It allows us to start training from scratch removing the need to have a initial set of labeled samples. In addition, it enables \qs to choose a diverse batch of samples in every cycle of active learning that allows a faster convergence consequently reducing the total cost of training. We have conducted an experiment to emphasize these points and quantitatively show how \daal results in a lower value of cost throughout the course of active learning. Fig.~\ref{fig:anneal} illustrates the results of this experiment. In this experiment, we tried three ways of obtaining the initial labeled set. (a) \emph{balanced}: An oracle selects a balanced limited set of data from a large unlabeled pool and annotates them.  (b) \emph{biased}: An oracle selects a biased unbalanced set of training data from a large unlabeled pool (c) \emph{beta} (our method): We select a batch with highest $q(x;\psi)$ score from the unlabeled pool. In our method, we used a large value of $\beta$ at the beginning ($\beta=4$ in this case) and used geometric annealing with some rate constant $\alpha$ ($\alpha=0.9$ in this case), i.e., we update the attention parameter $\beta\leftarrow0.9\times\beta$ at each AL cycle. A batch of size $32$ is queried by the \qs at each AL cycle. In this experiment we used $5000$ samples from MNIST dataset to train the VAE and put the remaining samples in the active learning pool.


\subsection{VAE latent space analysis} 
\begin{figure*}[h]
    \centering
    \begin{subfigure}[t]{0.24\textwidth}
        \centering
        \par\vfill
        \includegraphics[width=1.22\textwidth]{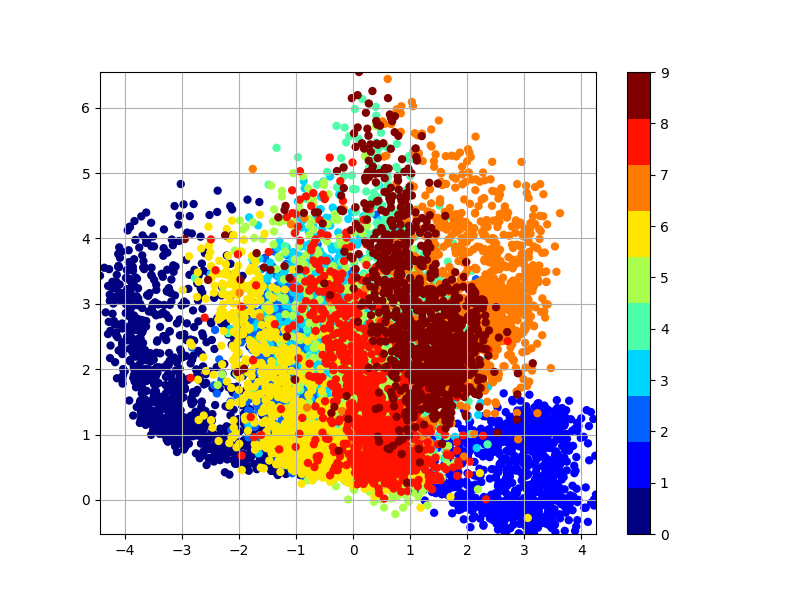}
        \subcaption{\small Ground-truth}
    \end{subfigure}%
    \foreach \n in {0, 1, 2}
    {
    \begin{subfigure}[t]{0.24\textwidth}
        \centering
        \includegraphics[width=1.22\textwidth]{figures/latent/candidates_latent_\n_before.png}
        \par\vfill
        \includegraphics[width=1.22\textwidth]{figures/latent/candidates_latent_\n_after.png}
        \subcaption{cycle $\n$}
    \end{subfigure}%
    }
    \caption{\small Each column depicts the query points at each cycle of AL in the latent space. Top row shows the predicted labels for each queried sample before the next AL cycle. Bottom row shows the predicted labels after training the classifier with the new queried samples. The ground-truth label of each point is shown on the leftmost figure for the reference.}
    \label{fig:mnist_latent}
\end{figure*}

We introduce this experiment to better investigate what actually happens while AL cycles are conducted by \daal when $\beta$ is annealed. We monitor the latent space of the employed VAE which is used by the \qs. Fig~\ref{fig:mnist_latent} illustrates the samples which are selected at each cycle of active learning in the latent space of VAE. The predicted label for each point by the classifier before and after each cycle is shown with corresponding colors. At each AL cycle, the chosen samples show a clustered distribution where each cluster corresponds to ambiguity in some parts of the decision boundary. For example, the decrease in the entropy of labels in locations $(2,0.5)$ at cycle 1 or $(-0.5,0.5)$ at cycle 2 is observed when we move from top row(\emph{before}) to bottom row(\emph{after}). This shows that \daal can reduce ambiguity of the decision boundaries for distant areas in the feature space. Conventional AL methods which are blind to structural information can reduce ambiguity of decision boundaries only on nearby locations or on one area at each cycle.

\section{Discussion}
We have proposed a simple modular method to augment the conventional active learning algorithms with the structural information of unsupervised data. We used variational autoencoder as a module that learns and encapsulates the structural information which is later used by the active learner to decide which samples are more informative. The flexibility and generality of this modular approach separates our work from the other studies who use some kind of structural information during active learning. Several experiments were done to enlighten different aspects of our proposed method. Synthetic and real datasets showed that our method is more robust against outliers. Even in the absence of outliers, having structural information enables the active learner to start with a less biased initial labeled set and take more diverse batches at each AL cycle. Future directions include combining other state-of-the-art generative~\citep{wang2016learning} and density estimator~\citep{deen} models with active training of discriminative models. Furthermore, generative components of the variational autoencoder and generative adversarial networks could enhance synthesized sampling strategies for active learning.


\bibliographystyle{unsrt}
\bibliography{arxiv}

\end{document}